\documentclass{article}
\usepackage{ijcai22}

\usepackage{times}

\usepackage{soul}
\usepackage{url}
\usepackage[hidelinks]{hyperref}
\usepackage[utf8]{inputenc}
\usepackage[small]{caption}
\usepackage{graphicx}
\usepackage{amsmath}
\usepackage{booktabs}
\usepackage{graphicx}
\usepackage{listings}
\usepackage{xcolor}
\definecolor{dkred}{rgb}{0.5,0,0}
\definecolor{dkgreen}{rgb}{0,0.6,0}
\definecolor{gray}{rgb}{0.5,0.5,0.5}
\definecolor{mauve}{rgb}{0.58,0,0.82}

\title{AutoVideo: An Automated Video Action Recognition System}

\author{
Daochen Zha$^1$\thanks{Those authors contribute equally to this project}
\and
Zaid Pervaiz Bhat$^2$\footnotemark[1] \and
Yi-Wei Chen$^2$\footnotemark[1] \and
Yicheng Wang$^2$\footnotemark[1] \and \\
Sirui Ding$^2$\footnotemark[1] \and
Jiaben Chen$^3$\footnotemark[1] \and
Kwei-Herng Lai$^1$\footnotemark[1] \and
Mohammad Qazim Bhat$^2$\footnotemark[1] \and \\
Anmoll Kumar Jain$^2$ \and
Alfredo Costilla Reyes $^1$ \and
Na Zou$^4$ \And
Xia Hu$^1$
\affiliations
$^1$ Department of Computer Science, Rice University \\
$^2$ Department of Computer Science and Engineering, Texas A\&M University \\
$^3$ School of Information Science and Technology, ShanghaiTech University \\
$^4$ Department of Engineering Technology and Industrial Distribution, Texas A\&M University \\
\emails
\{daochen.zha, khlai, acostillar, xia.hu\}@rice.edu, chenjb1@shanghaitech.edu.cn \\ \{zaid.bhat1234, yiwei\_chen, wangyc, siruiding, jain.anmollkumar, nzou1\}@tamu.edu 
}

\begin{document}

\maketitle

\begin{abstract}
Action recognition is an important task for video understanding with broad applications. However, developing an effective action recognition solution often requires extensive engineering efforts in building and testing different combinations of the modules and their hyperparameters. In this demo, we present AutoVideo, a Python system for automated video action recognition. AutoVideo is featured for 1) highly modular and extendable infrastructure following the standard pipeline language, 2) an exhaustive list of primitives for pipeline construction, 3) data-driven tuners to save the efforts of pipeline tuning, and 4) easy-to-use Graphical User Interface (GUI). AutoVideo is released under MIT license at \url{https://github.com/datamllab/autovideo}
\end{abstract}

\section{Introduction}
Action recognition is one of the most important tasks in video understanding~\cite{herath2017going}. Given a video clip, it aims to identify the human actions in the video, such as brush hair, cartwheel, catch, chew, clap, climb, etc. It has broad applications, such as security~\cite{meng2007human}, healthcare~\cite{gao2018human} and behavior analysis~\cite{poppe2010survey}.

Deep learning has achieved promising performance in action recognition~\cite{tran2015learning,wang2016temporal,carreira2017quo,zolfaghari2018eco,hou2019efficient}. However, developing a deep learning solution heavily relies on human efforts. First, we often need a very complex training pipeline, including but not limited to data loading, frame extraction, video cropping/scaling, video augmentation, model training, etc., which requires huge engineering efforts. Second, to achieve a good performance, a practitioner often needs extensive laborious trials on different combinations of the modules and their hyperparameters. 

\begin{figure}[t]
    \centering
    \includegraphics[width=0.72\linewidth]{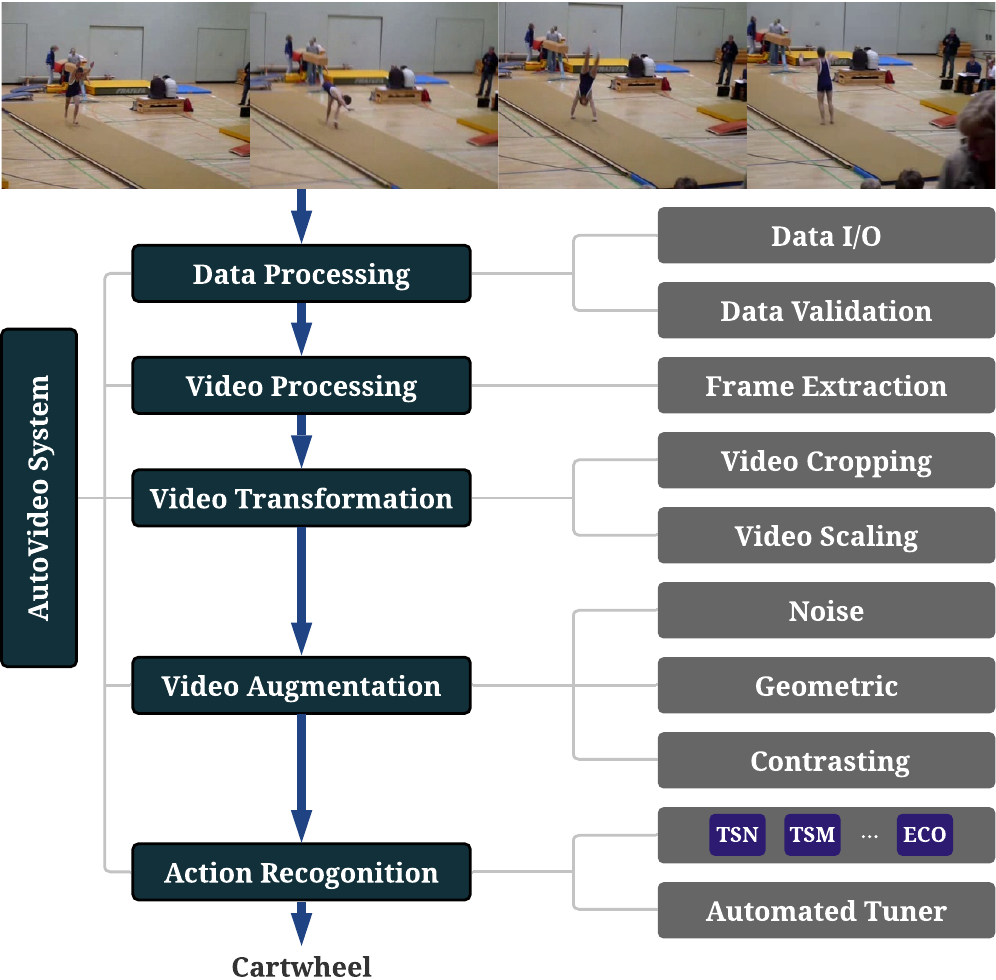}
    \caption{AutoVideo implements a complete training pipeline from data processing to action recognition, where each pipeline step can be instantiated with customized choices of primitives. Automated tuners are provided to help discover the best pipeline design.}
    \vspace{-10pt}
    \label{fig:overview}
\vspace{-5pt}
\end{figure}

To bridge this gap, in this demo, we present an automated video action recognition system named AutoVideo, which has several desirable features. \textbf{First, it is highly modular and extendable.} It is designed using the standard pipeline language under D3M infrastructure~\cite{milutinovic2020on,lai2021tods}, which defines primitives as basic building blocks and uses Directed Acyclic Graph~(DAG) to describe pipelines. The inputs, the outputs, and the hyperparameters of each module are well formatted so that we can easily develop and integrate new modules into the system with minimum engineering costs. \textbf{Second, it supports an exhaustive list of primitives for pipeline construction.} Specifically, we have so far supported 188 primitives from video processing to recognition models. Users can create various training pipelines via combining these primitives in different ways. \textbf{Third, it provides data-driven tuners to save the efforts of pipeline tuning.} We have implemented two commonly used AutoML tuners, including random search and Hyperopt~\cite{bergstra2013hyperopt}. They can automatically explore different primitive combinations and tune the hyperparameters to identify the best pipeline design. \textbf{Fourth, it provides a Graphical User Interface~(GUI).} In the GUI, users can manually construct pipelines in a drag-and-drop fashion, launch the training or search, and monitor the training progress.

While there are some other open-sourced video understanding libraries out there~\cite{mmaction2019,pytorchvideo2021}, they only provide a model zoo for users. In contrast, we provide an exhaustive list of primitives to allow users to create various pipelines or search with AutoML tuners. Unlike some other research-oriented AutoML studies for videos~\cite{piergiovanni2019evolving,assemblenet}, we target an easy-to-use and open-sourced package with unified interfaces for users. Thus, AutoVideo complements the existing efforts.

\begin{table}[]
 
  \centering
  \scriptsize
  \setlength{\tabcolsep}{12pt}
  \begin{tabular}{l|cc}
    \toprule
     Module & Number & Examples\\
    \midrule
    \midrule
    Data Processing & 5 & DatasetToDataFrame \\
    Video Processing & 1 & FrameExtraction \\
    Video Transformation & 3 & GroupScale, GroupCenterCrop  \\
    Video Augmentation & 170 & AdditiveGaussianNoise, Rotate \\
    Action Recognition & 9 & TSN, TSM, C3D, ECO  \\
    \midrule
    Total & 188 & - \\
    
  \bottomrule
\end{tabular}
\vspace{-10pt}
\caption{The number of primitives implemented in each module of AutoVideo with some example primitives.}
\label{tbl:primitives}
\vspace{-10pt}
\end{table}

\section{AutoVideo System}
Figure~\ref{fig:overview} shows an overview of the AutoVideo system. It implements a complete training pipeline consisting of data processing, video processing, video transformation, video augmentation, and action recognition, where each of the pipeline steps can be instantiated with customized choices of primitives, which leads to a large number of possible pipelines. Then two automated tuners are provided to automatically discover the best pipelines based on the performance. This section first introduces the standard pipeline language and then elaborates on our programming interface and GUI.


\subsection{Primitives and Pipelines}
We build AutoVideo upon D3M infrastructure~\cite{milutinovic2020on,lai2021tods}, which provides generic and extendable descriptions for modules and pipelines. The interface can accommodate various modules. Here, we provide an overview of the basic concepts. More details of the pipeline language can be found in~\cite{milutinovic2020on}.

\emph{Primitive} is the basic build block. It is an implementation of a function with some hyperparameters. A \emph{pipeline} is a Directed Acyclic Graph (DAG) consisting of several primitive steps. \emph{Data types}, such as DataFrame and NumPy Ndarrays, can be passed between steps in a pipeline. Each of the above concepts is associated with metadata to describe parameters, hyper-parameters, etc. Following this pipeline language, we wrap each component in AutoVideo (e.g., an action recognition algorithm or an augmentation module) as a primitive with some associated hyperparameters. We have so far supported 188 primitives, which are summarized in Table~\ref{tbl:primitives}, where data processing is designed for reading videos and the labels, video processing converts videos into frames, video transformation transforms the videos into the target size through cropping or scaling, video augmentation augments the videos in training, and action recognition trains deep learning models. Various pipelines can be constructed by combining these primitives in different ways. The genericity of the pipeline language allows AutoVideo to be easily extended to support other video-related tasks in the future.

\subsection{Programming Interface}
From the programming view, AutoVideo enables users to easily train any manually designed pipelines and also use automated tuners to help discover the best pipelines. A minimum example of fitting a pipeline is given below.

\begin{figure*}[t]
  \centering
    \includegraphics[width=1.0\textwidth]{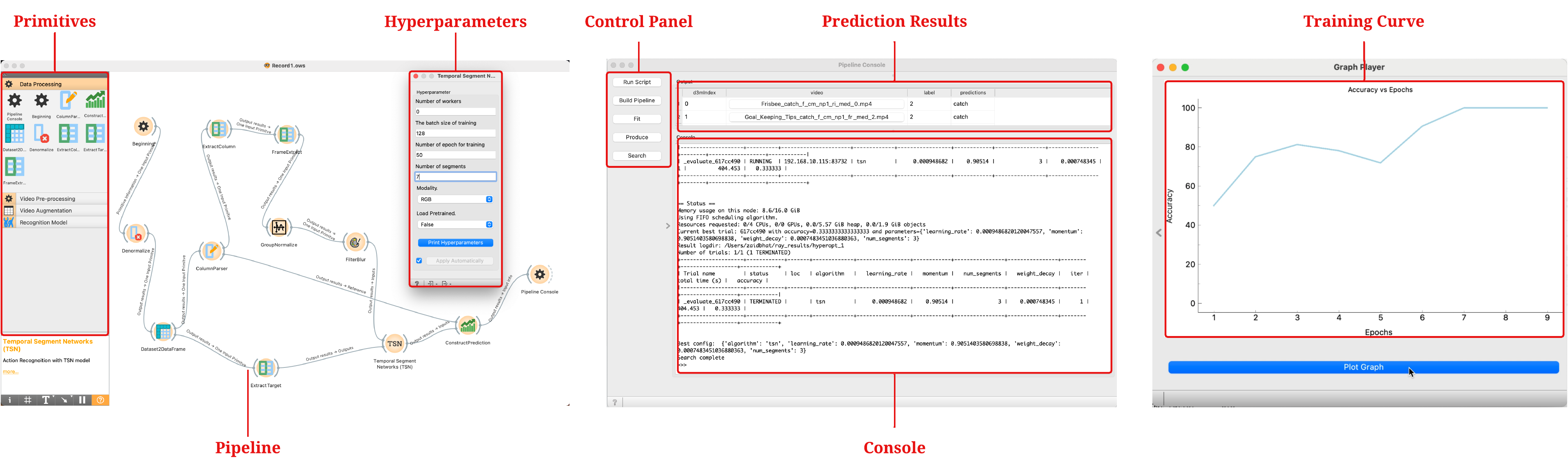}
    \vspace{-14pt}
  \caption{GUI of AutoVideo to support pipeline construction in a drag-and-drop fashion (left), fitting/evaluating/searching pipelines (middle), and training progress visualization (right).}
  \vspace{-12pt}
  \label{fig:gui}
\end{figure*}

\begin{lstlisting}[title={Example 1: Code snippet of fitting a model.},captionpos=b]
from autovideo import fit, build_pipeline

# Build pipeline based on config
pipeline = build_pipeline({
    "transformation": [("Scale", {"scale_size": (128,128)})],
    "augmentation": [
        ("arithmetic_AdditiveGaussianNoise", {"scale": (0, 0.2*255)}),
        ("geometric_Rotate", {"rotate": (-45, 45)}),
        ("contrast_LogContrast", {"gain": (0.6, 1.4)}),
    ],
    "multi_aug": "meta_Sometimes",
    "algorithm": "tsn",
    "load_pretrained": False,
    "epochs": 50,
})

# Fit
fit(train_dataset, train_media_dir, target_index=target_index, pipeline=pipeline)
\end{lstlisting}
\vspace{-5pt}
\texttt{train\_dataset} is the DataFrame describing the video file names and labels, \texttt{train\_media\_dir} contains the directory of the video files, and \texttt{target\_index} specifies which column is label. Pipelines are described with a configuration dictionary specifying the primitives and hyperparameters in each pipeline step. Users can customize the configuration in \texttt{build\_pipeline} to build pipelines with different combinations of the primitives and the hyperparameters. The \texttt{fit} function will return the fitted pipeline that can be saved for making predictions. However, manually tuning the pipeline is laborious and tedious. Thus, we can alternatively use automated tuners. An example is presented below.

\begin{lstlisting}[title={Example 2: Code snippet of automated tuning.},captionpos=b]
from ray import tune
from autovideo.searcher import RaySearcher

search_space = {
    "augmentation": {
        "aug_0": tune.choice([
            ("arithmetic_AdditiveGaussianNoise",),
            ("arithmetic_AdditiveLaplaceNoise",),
        ]),
        "aug_1": tune.choice([
            ("geometric_Rotate",),
            ("geometric_Jigsaw",),
        ]),
    },
    "multi_aug": tune.choice([
        "meta_Sometimes",
        "meta_Sequential",
    ]),
    "algorithm": tune.choice(["tsn"]),
    "learning_rate": tune.uniform(0.0001, 0.001),
    "momentum": tune.uniform(0.9,0.99),
    "weight_decay": tune.uniform(5e-4,1e-3),
    "num_segments": tune.choice([8,16,32]),
}

config = {
    "searching_algorithm": "hyperopt",
    "num_samples": 100,
}

searcher = RaySearcher(train_dataset, train_media_dir, valid_dataset=valid_dataset, valid_media_dir=valid_media_dir)

best_config = searcher.search(
    search_space=search_space,
    config=config
)
\end{lstlisting}

Here, \texttt{valid\_dataset} is the DataFrame associated with the validation data. \texttt{search\_space} is a dictionary specifying the augmentation methods, recognition models, and the ranges of hyperparameters the tuner will cover. We follow the search space definitions in Ray~\cite{moritz2018ray}, which allows the users to flexibly design the search space. \texttt{config} specifies some tuning configurations, such as the type of the tuner and the search budget. The automated tuner will search for the best pipeline designs within the search space and return the best-discovered pipeline configuration in \texttt{best\_config}, which can be then used to re-fit the pipeline.

\subsection{GUI}

Figure~\ref{fig:gui} presents our GUI to help users construct pipelines, fit the constructed pipelines, and make predictions. The GUI is implemented based on Orange~\cite{demvsar2004orange}.

The left-hand side of Figure~\ref{fig:gui} illustrates the canvas for pipeline building. Users can construct a pipeline in a drag-and-drop fashion by dragging icons (primitives in the left) to the canvas and connecting them with lines. For each of the primitives, we can double-click the icon to modify the default hyperparameters of the primitive. Once a pipeline is built, the GUI will convert the constructed pipeline into the configuration dictionary as in Example 1 and the backend will instantiate the pipeline. In addition, users can save a pipeline and load it later just like editing a document in Microsoft Word.

The middle and right-hand sides of Figure~\ref{fig:gui} show the pop-up windows for pipeline fitting and training progress monitoring, respectively. Specifically, the users can fit the pipeline, make predictions with a fitted pipeline, or perform automated searching by simply clicking a button in the control panel. For example, by clicking the ``fit'' button, the backend will train the model and visualize the training curve once the training is finished. By clicking the ``produce'' button, the backend will make predictions on the testing data. The ``search`` button will launch the automated tuners for pipeline search and save the best pipeline. The prediction results will be displayed at the top of the window, where the users can watch each of the testing videos and check its predictions and labels. In addition, we also provide an interactive console, where the users can interact with the backend with command lines. This design provides flexibility to meet different needs of the users.

\section{Effectiveness of Pipeline Search}
We conduct a preliminary experiment to showcase the effectiveness of pipeline search. We define a search space that allows choosing 3 augmentation primitives from the categories of \texttt{arithmetic}, \texttt{geometric}, and \texttt{color}, and tuning the hyperparameters of the TSN model with learning rate [0.001,0.0001], momentum [0.9,0.99], weight decay [5e-4,1e-3], number of segments {8,16,32}. We separate 5\% of the training data for validation purposes. We apply the random search and Hyperopt tuners to search the pipeline configurations. We run the search with 50 samples and evaluate the best configuration discovered in the search.

In Table~\ref{tbl:perf}, we compare the two tuners as well as the default pipeline configuration on HMDB-51\footnote{ \url{https://serre-lab.clps.brown.edu/resource/hmdb-a-large-human-motion-database/}}, a task that aims to identify 51 human actions, and a subset of it with the first six actions named HMDB-6. Both the tuners outperform the default pipeline by a large margin, which verifies the effectiveness of pipeline search and hyperparameter tuning. We defer a more comprehensive evaluation to our future work since it is out of the scope of our demo presentation.




\begin{table}[]
  \centering
  \scriptsize
  \setlength{\tabcolsep}{10pt}
  \begin{tabular}{l|cc}
    \toprule
      & HMDB-6 & HMDB-51\\
    \midrule
    \midrule
    Default Pipeline & 80.00\% & 34.84\% \\
    Random Search & 94.8\% & 51.24\% \\
    Hyperopt & 94.5\% & 54.71\% \\
  \bottomrule
\end{tabular}
  \vspace{-8pt}
  \caption{Accuracy of tuners in AutoVideo.}
  \label{tbl:perf}
\vspace{-15pt}
\end{table}

\section{Conclusions and Future Work}
We present AutoVideo, an automated video action recognition system. It provides a highly modular implementation of 188 primitives, on which users can flexibly create pipelines. It also supports automated tuners and an easy-to-use GUI to help researchers/practitioners develop prototypes. We will actively maintain AutoVideo and develop more features based on our previous research. We plan to wrap more primitives beyond action recognition from our research codes in object detection~\cite{wang2022bed} and outlier detection~\cite{zha2020meta,li2020pyodds,lai2021revisiting}, and tuners with reinforcement learning~\cite{zha2021douzero,zha2021rlcard,zha2019experience,zha2020rank}.

\bibliographystyle{named}
\bibliography{ref}

\end{document}